\documentclass{article}
\usepackage{arxiv}
\usepackage[utf8]{inputenc} % allow utf-8 input
\usepackage[T1]{fontenc}    % use 8-bit T1 fonts
\usepackage{hyperref}       % hyperlinks
\usepackage{url}            % simple URL typesetting
\usepackage{booktabs}       % professional-quality tables
\usepackage{amsfonts}       % blackboard math symbols
\usepackage{nicefrac}       % compact symbols for 1/2, etc.
\usepackage{microtype}      % microtypography
\usepackage{lipsum}

\usepackage[ruled,vlined]{algorithm2e}
\usepackage{authblk}
\usepackage{graphicx}

\title{Deep interpretable architecture for plant diseases classification}

\author[1,2,3]{Mohammed Brahimi}
\author[2]{Sa\"{i}d Mahmoudi}
\author[1]{Kamel Boukhalfa}
\author[4]{Abdelouhab Moussaoui}
\affil[1]{Computer Science Department, USTHB University, Algiers, Algeria 
\authorcr Email: {\tt kboukhalfa@usthb.dz}}
\affil[2]{Computer science department, Faculty of engineering, University of Mons, Belgium \authorcr Email: {\tt Said.MAHMOUDI@umons.ac.be}}
\affil[3]{Computer Science Department, Mohamed El Bachir El Ibrahimi University, Bordj Bou Arreridj, Algeria \authorcr Email: {\tt m\_brahimi@esi.dz}}
\affil[4]{Department of Computer Science, Setif 1 University, Setif, Algeria \authorcr Email: {\tt moussaoui.abdel@gmail.com} }

\begin{document}
\maketitle

\begin{abstract}
Recently, many works have been inspired by the success of deep learning in computer vision for plant diseases classification. Unfortunately, these end-to-end deep classifiers lack transparency which can limit their adoption in practice. In this paper, we propose a new trainable visualization method for plant diseases classification based on a Convolutional Neural Network (CNN) architecture composed of two deep classifiers. The first one is named Teacher and the second one Student. This architecture leverages the multitask learning to train the Teac1her and the Student jointly. Then, the communicated representation between the Teacher and the Student is used as a proxy to visualize the most important image regions for classification. This new architecture produces sharper visualization than the existing methods in plant diseases context. All experiments are achieved on PlantVillage dataset that contains 54306 plant images.  

\end{abstract}

% keywords can be removed

\section{Introduction}
Plant diseases cause great damages to agriculture crops by significantly decreasing production \cite{HANSSEN201231}. Protecting plants from diseases is vital to guarantee the quality and the quantity of crops \cite{Brahimi:2017}. A successful protection strategy  starts with an early detection of the disease and the right treatment to prevent its spreading. Many studies proposed the use of Convolutional Neural Network (CNN) to detect and classify diseases. This new trend produced more accurate classifiers compared to shallow machine learning approaches based on hand crafted features \cite{Fujita:2016,Kawasaki:2015,Brahimi:2017,Nachtigall:2016}.  Despite all these successes, CNN still suffers from the lack of transparency that limits its spreading in many domains. These CNNs are complex deep models that yield good results at the expense of explainability and interpretability. High accuracy is not sufficient for plant diseases classification. Users also need to be informed how the detection is achieved and which symptoms are present in the plant.

In this paper, we propose a classification and visualization architecture based on multitask learning of two classifiers: Teacher and Student. This architecture represents a trainable visualization method for plant diseases classification. The main contribution is the design of an interpretable deep architecture able to do classification and visualization simultaneously. The visualization algorithm is embedded directly in the network design instead of using it after the training as post treatment.

\section{Related works}
Visualization algorithms are used to explain CNN decision using a heatmap. This heatmap highlights the importance of each pixel for the classifier's decision. These methods backpropagate to the input image the discriminant features used by the network for classification. Most of these methods use heuristics during the backpropagation. For example, gradients based methods like deconvolution \cite{Zeiler:13} and guided backpropagation \cite{Springenberg:2015} filter some signals during the backpropagation according to heuristic rules. These filtered signals are ignored to produce sharp visualizations. Furthermore, Layer Wise Relevance Propagation (LRP) methods \cite{Bach:2015,MONTAVON:2017} are based on choosing the layer's roots to determine propagation rules for each layer in the network. The global visualization method GRAD-CAM \cite{Selvaraju:2017} projects the features from the penultimate convolution layer based on a linear interpolation which degrades the precision of the produced heatmaps. These visualization algorithms can produce different heatmaps according to the chosen heuristics and propagation rules, which makes the understanding of the classifier difficult. In the present paper, we combine two classifiers to extract discriminant features from images. These discriminant features are extracted by the first classifier and projected as an input for the second classifier. This trainable visualization is similar to segmentation architectures like U-Net \cite{Ronneberger:2015} where the supervision masks are replaced with a second classifier.

\section{Proposed method}
\begin{figure}[htbp]
%\centering
\centerline{\includegraphics[scale=0.7]{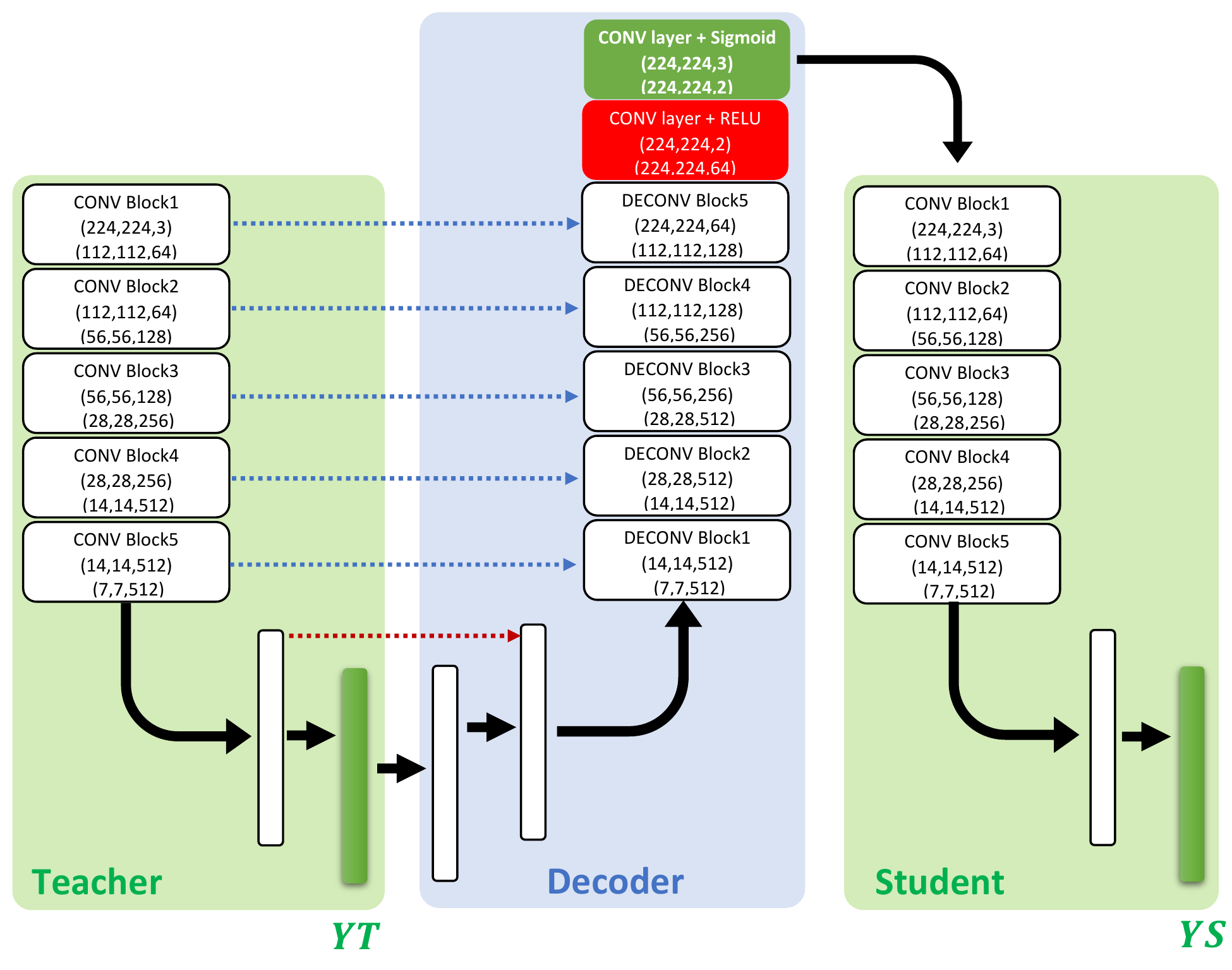}}
\caption{Teacher/Student network architecture.}
\label{fig:arch}
\end{figure}

We propose a classification and visualization generic architecture, named \textbf{Teacher/Student} architecture, based on learning transfer from a first classifier (Teacher) to a second one (Student). This learning transfer from the Teacher to the Student is achieved using an autoencoder having the Teacher as an encoder. The decoder consumes the Teacher latent representations to reconstruct an image with the same dimension of the input image. This image is used as an input of the Student. The whole network (Teacher + Decoder + Student) is trained to minimize jointly the losses of the two classifiers (Teacher and Student). More formally, the network has two outputs $YT$ (Teacher output) and $YS$ (Student output). During the training, the loss function (\ref{eq:Loss}) is minimised. The hyperparameter $ \alpha $ represents the tradeoff between the Teacher loss (\ref{eq:LossTeacher}) and the Student loss (\ref{eq:LossStudent}). 

\begin{equation}\label{eq:Loss}
Loss = \alpha * LossTeacher + (1-\alpha)*LossStudent 
\end{equation}

\begin{equation}\label{eq:LossTeacher}
LossTeacher = -\frac{1}{N}\sum_{i=1}^N\sum _{j=1}^C Y_{j}^{i}log(YT_{j}^{i}) 
\end{equation}

\begin{equation}\label{eq:LossStudent}
LossStudent = -\frac{1}{N}\sum_{i=1}^N\sum _{j=1}^C Y_{j}^{i}log(YS_{j}^{i}) 
\end{equation}

The Teacher/Student network is designed to reconstruct an image containing the discriminant features formed by the Teacher to help the Student training. As a side effect of this design, the reconstructed image can be used as a visualization of the important regions for the classification. This architecture represents an autoencoder to denoise the image from irrelevant features from the classification viewpoint. The difference between the usual denoising autoencoder and this architecture lies in the loss function design. The denoising autoencoder minimizes a reconstruction loss  while this architecture minimizes the classification loss of two classifiers. This proposed architecture  is also designed to extract the important regions for classification without using masks. For this reason, any segmentation architecture can be modified to fit the proposed architecture by modifying the loss function to include a Student classifier and avoid the segmentation masks.

Fig.~\ref{fig:arch} details the Teacher/Student architecture. For the sake of simplicity,  VGG16 \cite{Simonyan:2015} architecture is used as Teacher and Student. Nevertheless, the Teacher/Student architecture is flexible and other classification architectures can be used as Teacher or Student. To use another architecture, the decoder must be adapted to inverse the Teacher's layers to reconstruct the input of the Student.

The Teacher/Student architecture is composed of the following components:
\subsection{Teacher/Student architecture}
The Teacher and the Student architectures are identical to standard architecture VGG16 \cite{Simonyan:2015}. Skip connections (blue arrows) are used from the Teacher to the decoder. This skip connections concatenate the input tensors of pooling layer of each convolution block  with deconvolution block tensors. %Student architecture is used only to train the decoder and can be pruned at the end of training to optimize the network size. 
\subsection{Reversed fully connected layers}
Convolutional autoencoders reverse only convolution layers to do reconstruction task.  Here, the decoder requires discriminant features from fully connected layers. Therefore, two fully connected layers are used to reverse the Teacher's fully connected layers. Furthermore, a skip connection (red arrow) is used to reinforce the decoder by adding the vector of the Teacher's first fully connected layer.    
\subsection{Deconvolution blocks}

\begin{table}[htbp]
  \centering
  \caption{Deconvolution blocks details.}
    \begin{tabular}{|c|c|c|}
    \hline
    \textbf{Layer} & \textbf{Input tensor} & \textbf{Output tensor} \\
    \hline
    Upsampling2d  & $(X,Y,Z)$ & $(2X,2Y,Z)$ \\
    \hline
    Conv2D & $(2X,2Y,Z)$ & $(2X,2Y,Z1)$ \\
    \hline
    Concatenate & $(2X,2Y,Z1) + (2X,2Y,Z1)$ & $(2X,2Y,2Z1)$ \\
    \hline
    Conv2D & $(2X,2Y,2Z1)$ & $(2X,2Y,Z1)$ \\
    \hline
    Conv2D & $(2X,2Y,Z1)$ & $(2X,2Y,Z1)$ \\
    \hline
    \end{tabular}%
  \label{tab:deconvBlock}%
\end{table}%

Deconvolution blocks reverse the flow of tensors to form the reconstructed image. The details of this block are shown in Tab.~\ref{tab:deconvBlock}. This block consumes an input tensor  of dimension $(x,y,z)$ and produce a tensor of dimension $( 2x,2y,z_1)$. In each deconvolution block, tensor is upsampled, concatenated with the corresponding tensor of skip connection and the depth of resulted tensor after concatenation is reduced. The first two deconvolution block (DECONV Block1, DECONV Block2) double the input tensor spatially without changing its depth $(z_1=z)$. However, the other deconvolution blocks double the tensor spatially and reduce its depth $(z_1=z/2)$.
\subsection{Reconstructed image refinement}
After many stages of deconvolution, the spatial dimension of the produced tensor will be equal to input image spatial dimension. However, the depth of the tensor must be reduced to match the depth of the input image. To do this, two convolution layers are applied. The first one (red in Fig.~\ref{fig:arch}) reduces the depth to two channels which applies pressure on the data flow by throwing unnecessary details. The second convolution layer (green in Fig.~\ref{fig:arch}) expands the tensor to three channels to use it as the Student's input. The last decoder's convolution layer uses sigmoid activation function to scale the values in the interval $[0,1]$. This final tensor is used as Student's input and can be used also as proxy to understand the communication between the Teacher and the Student.  

\section{Experimental results}
Experimental results are conducted  using the segmented version  of PlantVillage dataset with black backgrounds \cite{David:2015}. This dataset includes 54306 images of 14 crop species with 38 classes of diseases and healthy plants. The data set is split into a training set that contains 32572 images and  a validation set that contains 21734 images. The code source of the proposed architecture is available at \textit{\url{https://github.com/Tahedi1/Teacher\_Student\_Architecture}}.

\subsection{Classification results}

\begin{figure}[htbp]
\centerline{\includegraphics[scale=0.7]{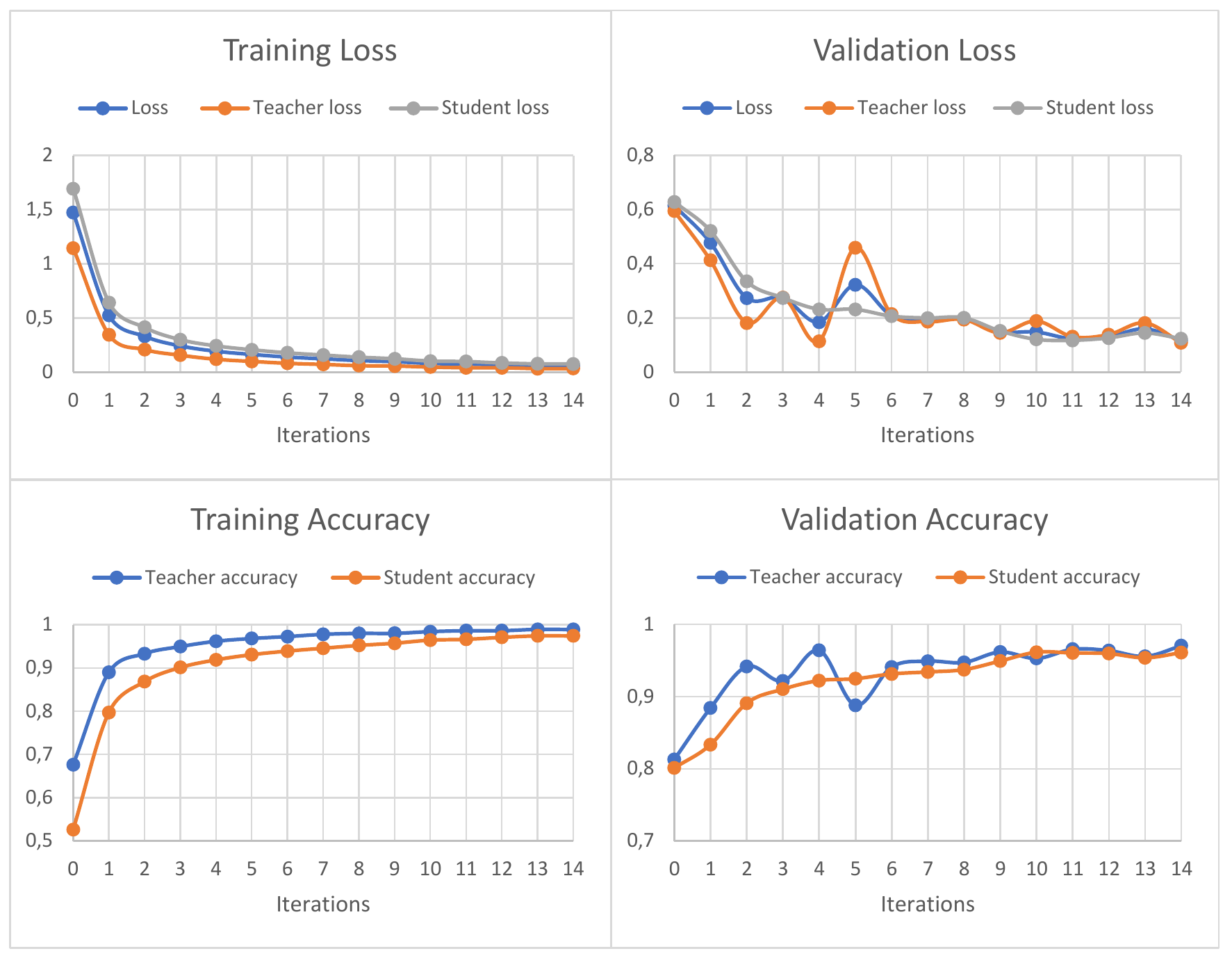}}
\caption{Classification results of the Teacher/Student architecture.}
\label{fig}
\end{figure}

In this section, we present the results of training and validation of the proposed architecture. The training is based on gradient descent algorithm with the following hyperparameters: (learning rate $=1e-4$, momentum $=0.9$, batch size= 16, number of iterations $= 15$, multitask hyperparamater  $\alpha =0.4$).  Training and validation are executed on a workstation containing Graphical Processing Unit GPU Nvidia GTX 1080.

At the beginning of training, the Teacher is more accurate than the student. This may be explained by the dependency of the Student on the representation constructed by the Teacher. However, the loss and the accuracy of the Teacher and the Student converge at the end of the training because the communicated representation  becomes stable.
Furthermore,  the Student's loss is more stable than the Teacher's loss in validation which reinforce the hypothesis of the transfer learning from the Teacher to the Student. This transfer learning is achieved through the quality of the reconstructed image. This reconstructed image focuses on discriminant regions and filters non-discriminant regions. To assess this information filtering mechanism, we analyze the architecture as a visualization method in the following sections.
\subsection{Visualization results}
\begin{figure}[htbp]
\centerline{\includegraphics[scale=0.8]{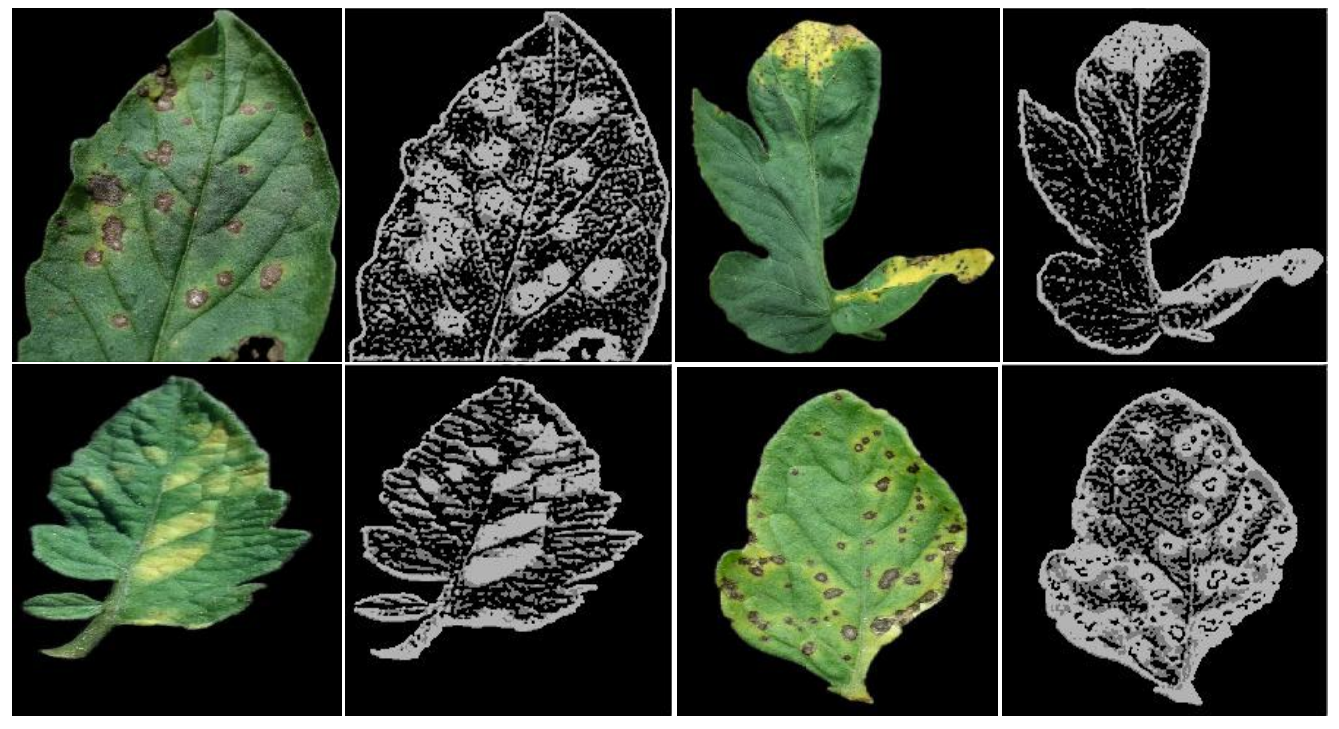}}
\caption{Visualization images of the Teacher/Student architecture.}
\label{fig:visio1}
\end{figure}
\begin{figure}[htbp]
\centerline{\includegraphics[scale=0.8]{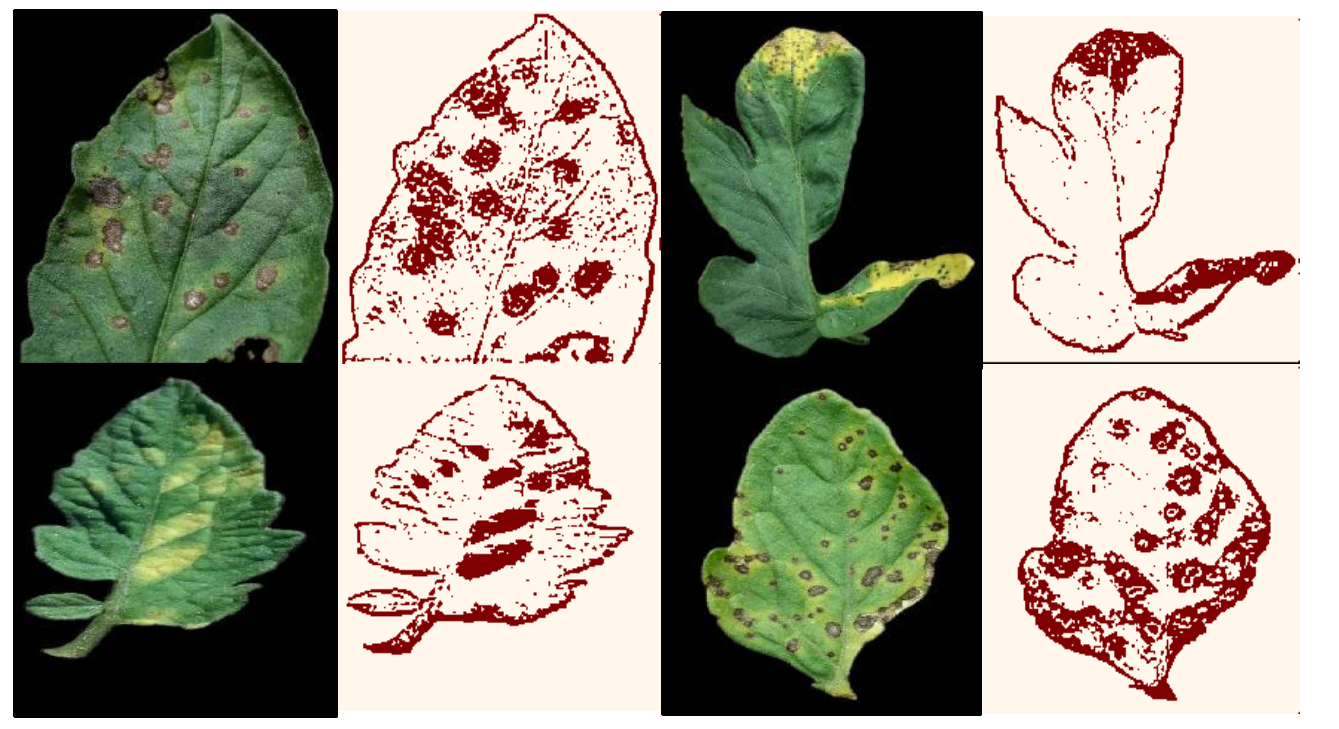}}
\caption{Heatmaps after threshoulding the visualizations images.}
\label{fig:visio2}
\end{figure}

The visualizations depicted on  Fig.~\ref{fig:visio1} represent the reconstructed three channels images used as an input for the Student. In Fig.~\ref{fig:visio2}, important regions are segmented using a simple binary thresholding algorithm (threshold  = 0.9). The thresholding algorithm is applied after a simple aggregation across channels of reconstructed image (noted $V$) to produce one channel heatmap. To produce this heatmap from the reconstructed image $V$, the formula (\ref{eq:aggr}) is applied. This formula measures the distance between pixel's color and black $(0,0,0)$.   
\begin{equation}\label{eq:aggr}
Heatmap(i,j) = \sqrt{V(i,j,0)^2 +V(i,j,1)^2+V(i,j,2)^2}    
\end{equation}

The produced heatmaps (Fig.~\ref{fig:visio2}) show clearly the symptoms of the  plant disease. Furthermore, these heatmaps are sharp and precise. The healthy regions of the leaves are filtered and only the important regions are highlighted. In the next section, the proposed method is compared quantitatively to other methods using perturbation curves to show its effectiveness.
\subsection{Comparison with visualization algorithms}
In this section, the proposed method is compared to the following visualization algorithms : visualization based on gradient \cite{Simonyan:2014}, Grad-CAM \cite{Selvaraju:2017} and Layer-wise Relevance Propagation (LRP) methods (Deep Taylor \cite{MONTAVON:2017}, LRP-Epsilon \cite{Bach:2015}, LRP-Z \cite{Bach:2015}). 

\subsubsection{Heatmaps comparison} 
\begin{figure}[htbp]
\centerline{\includegraphics[]{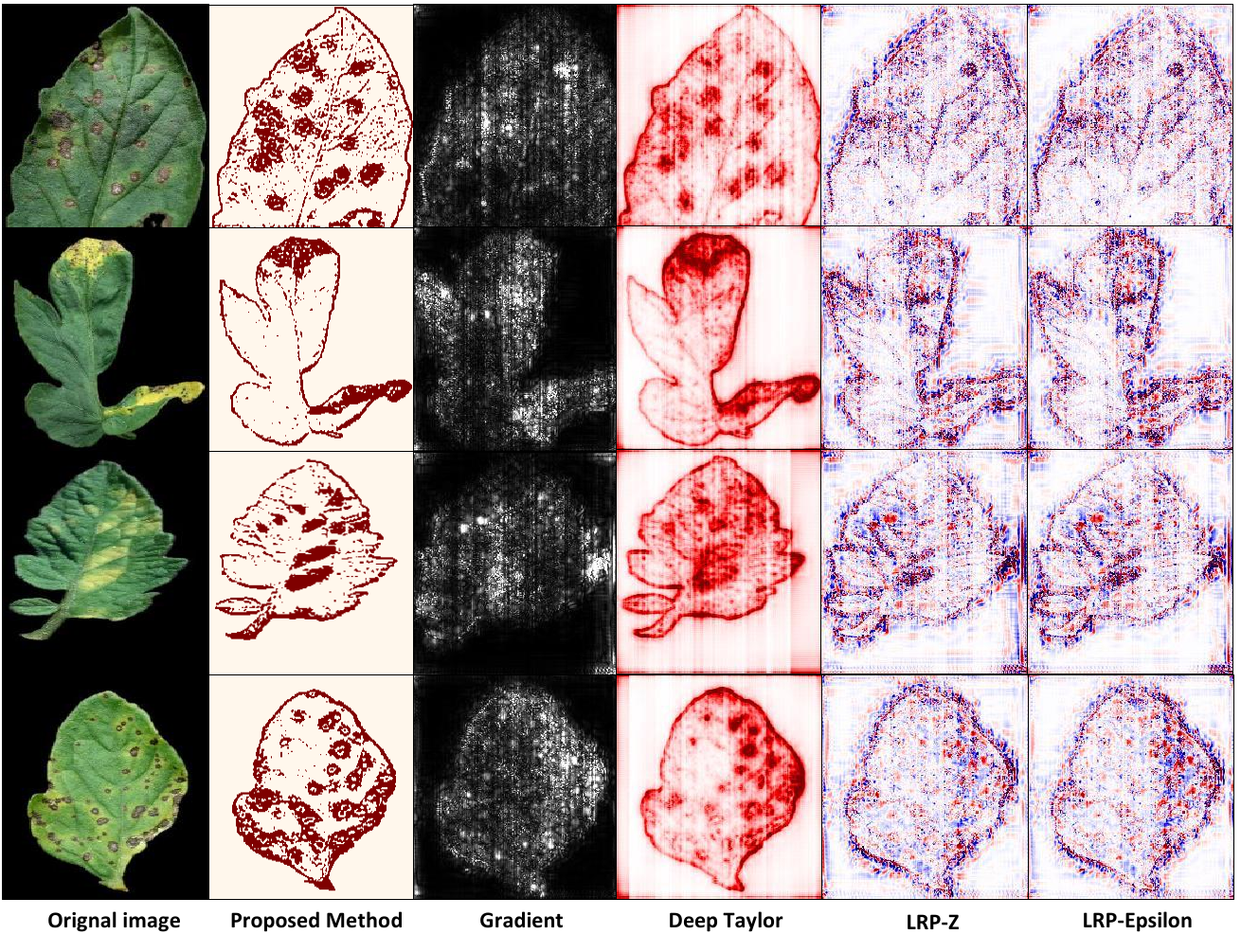}}
\caption{Comparison with visualization algorithms.}
\label{fig:visio3}
\end{figure} 

\begin{figure}[htbp]
\centerline{\includegraphics[width=\linewidth]{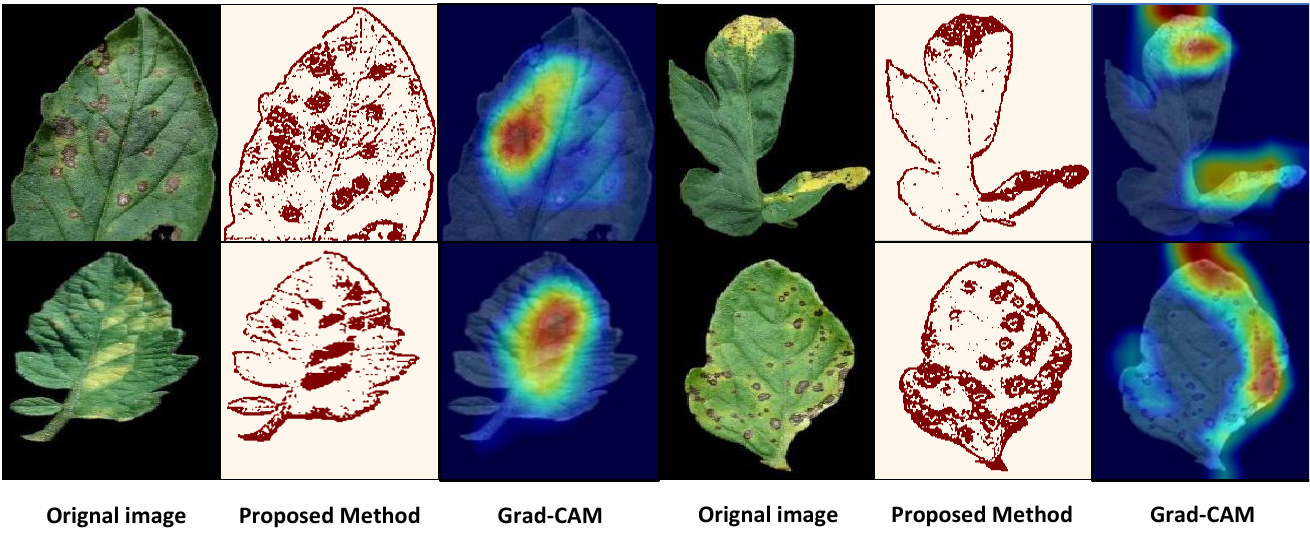}}
\caption{Comparison with Grad-CAM.}
\label{fig:visio4}
\end{figure} 

Fig.~\ref{fig:visio3} shows the difference between the heatmaps of the visualization algorithms. The proposed algorithm's heatmaps are sharper than the other heatmaps. Indeed, gradient, LRP-Z and LRP-Epsilon heatmaps are noisy and difficult to explain. The gradient heatmaps are noisy because the gradients measure the pixel's sensitivities instead of their contributions. Beside, the presence of negative and positive contributions in LRP heatmaps makes them difficult to understand. On the other hand, Deep Taylor algorithm has  good and clear heatmaps compared to other algorithms. The heatmaps of Deep Taylor algorithm highlight almost the same highlighted regions by our algorithm, but with some activated regions on the background. In the proposed algorithm, the background is completely deactivated which gives clean heatmaps.

Fig.~\ref{fig:visio4} shows the difference between the proposed method and Grad-CAM. The Grad-cam algorithm localizes globally  the important regions. Furthermore, the Grad-CAM visualizations miss some important regions highlighted by the proposed method. This inaccurate localization is due to the resizing used by Grad-CAM to propagate the contributions from the last convolution layer to the input image. In the Teacher/Student architecture, this propagation is ensured by a trainable decoder which makes the visualizations more precise.

\subsubsection{Histograms of heatmaps values}
all the produced heatmaps are normalized in the interval $[0,1]$ to analyze the distribution of values in each method. This normalization based on equation (\ref{eq:Normal}) where $Min$ and $Max$ are the minimum and the maximum of $Heatmap$ respectively.

\begin{equation}\label{eq:Normal}
Heatmap(i,j)=\frac{Heatmap(i,j)-Min}{Max-Min} 
\end{equation}

\begin{figure}[htbp]
\centerline{\includegraphics{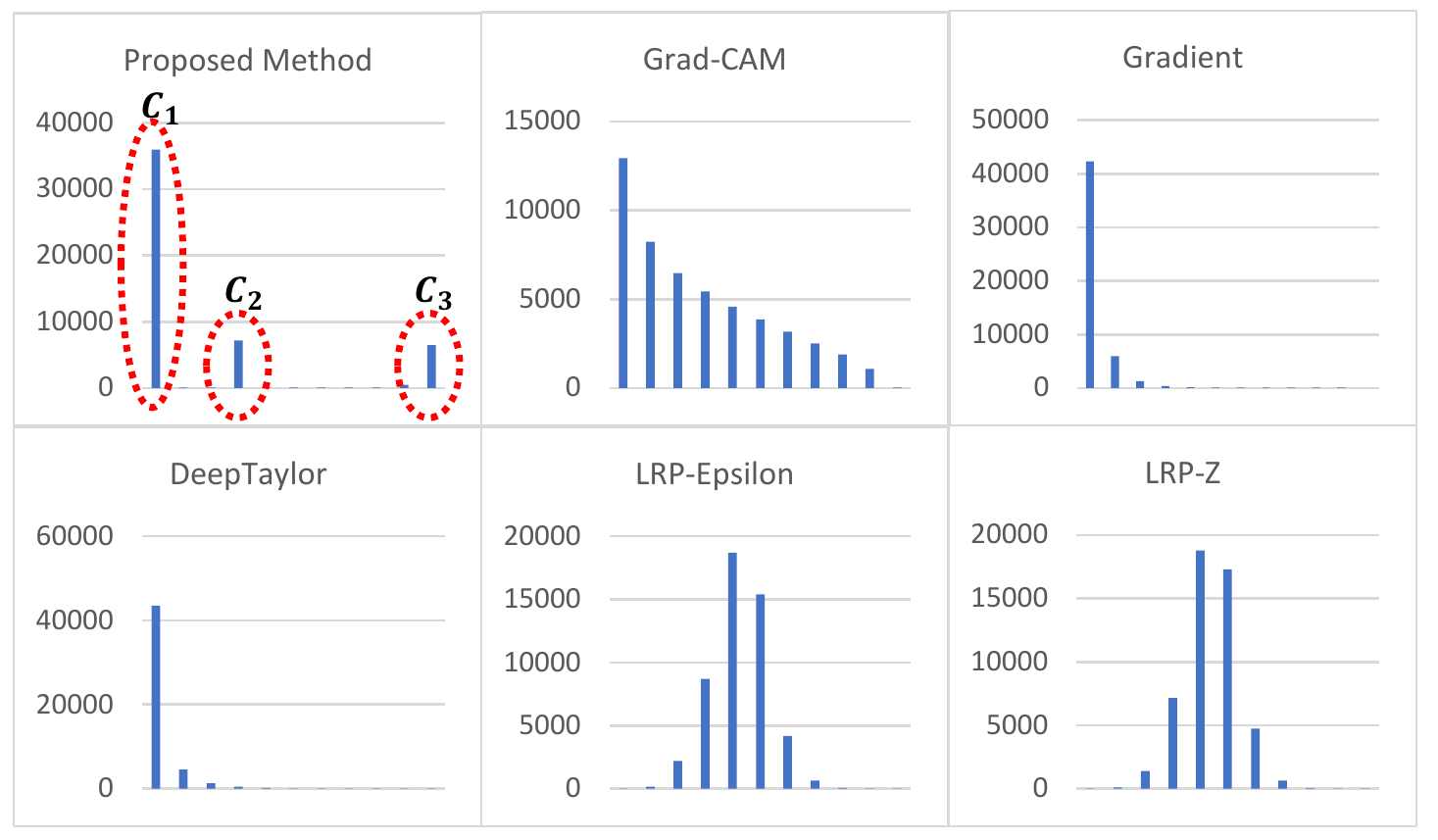}}
\caption{Histograms of the visualization algorithms.}
\label{fig:hist}
\end{figure}

Histograms of  methods are shown in Fig.~\ref{fig:hist}. We notice that the heatmaps values distribution of the proposed method is different from the other methods. Grad-CAM, gradient and Deep Taylor distributions are concentrated in very small values while the density decreases gradually as values increase. LRP-Epsilon and LRP-Z have gaussian-like distribution centered on 0.5.  In contrast, the proposed method distribution has three clusters: 
\begin{itemize}
\item Cluster1 (\(C_1)\): $72 \%$  of deactivated pixels  where the heatmap  values are zeros  ($ Heatmap(pixel) = 0$). This cluster represents the background and non-discriminant pixels.  
\item Cluster2 (\(C_2)\): $ 15 \% $ of pixels having small heatmap's values ($ 0.2 < Heatmap(pixel) \leq 0.3 $).
\item Cluster3 (\(C_3)\): $ 13 \% $ of pixels having high heatmap values ($ 0.9 < Heatmap(pixel) \leq 1$) and can be considered as important pixels for the classifier.
\end{itemize}

\subsubsection{Perturbation curves}

\begin{algorithm}
 \DontPrintSemicolon

 \KwIn{$X$ : Input image, \\
 \hspace{1.1cm}$Heatmap$ : Heatmap of input image $X$}
 
 \KwOut{$PC$ : Points list of the perturbation curve,\\
 \hspace{1.3cm}$AOCP$: Area over perturbation curve }
 
 $X^0 \leftarrow X$ 
 
 $PC.Append((0,f(X^0)))$  
 
 $AOPC \leftarrow 0 $
 
 $B \leftarrow  1 $ 
 
 $j \leftarrow 1$ 
 
 \For{$j\gets1$ \KwTo $11$ }
 {
    \If{$j<11$}
    {
        $R \leftarrow \{pixel : B-\frac{1}{10} < Heatmap(pixel) \leq B \} $
    }
    \Else
    {
        $R \leftarrow \{pixel : Heatmap(pixel) = 0 \}$
    }
    $X^{j} \leftarrow Erase(X^{j-1},R)$ 
    
    $PC.Append((j,f(X^{j})))$
    
    $AOPC \leftarrow AOPC + (f(X^0)-f(X^{j}))$ 
    
    $ B \leftarrow B - \frac{1}{10}$
 }
 $AOPC \leftarrow \frac{AOPC}{11}$ 
 
 \Return $PC,AOPC$ 
 
\caption{Perturbation curve experiment for one image $X$.}
\label{Algo:1}
\end{algorithm}

To measure the quality of visualization method quantitatively, the produced heatmap is considered as a ranking function of pixels. Good heatmap ranks the pixels correctly according to their importance for the classification. Therefore, if we start erasing the pixels having high values in the heatmap then the classifier output decreases rapidly. To evaluate this ranking, the heatmap values are discretized into intervals of size $\frac{1}{10}$. Afterwards, pixels are erased iteratively according to their heatmaps values in descending order. This erasing procedure is formulated in Algorithm.1%~\ref{Algo:1}. 
To erase one pixel, a small black square with a size of three pixels centered on the pixel of interest is used.  All dataset images have a black background which motivates the use of black color as reference to erase with. The function $f(X^j)$ gives the classifier estimation of the probability that $X^j$ has the same class of initial image $X^0=X$. The perturbation curve is traced using points $(j,f(X^{(j)}))$ to track the evolution of the classifier output during the erasing. To produce this perturbation curve that characterizes a visualization method, the perturbation curves of validation images  are averaged. 

\begin{figure}[htbp]
\centerline{\includegraphics[scale=0.8]{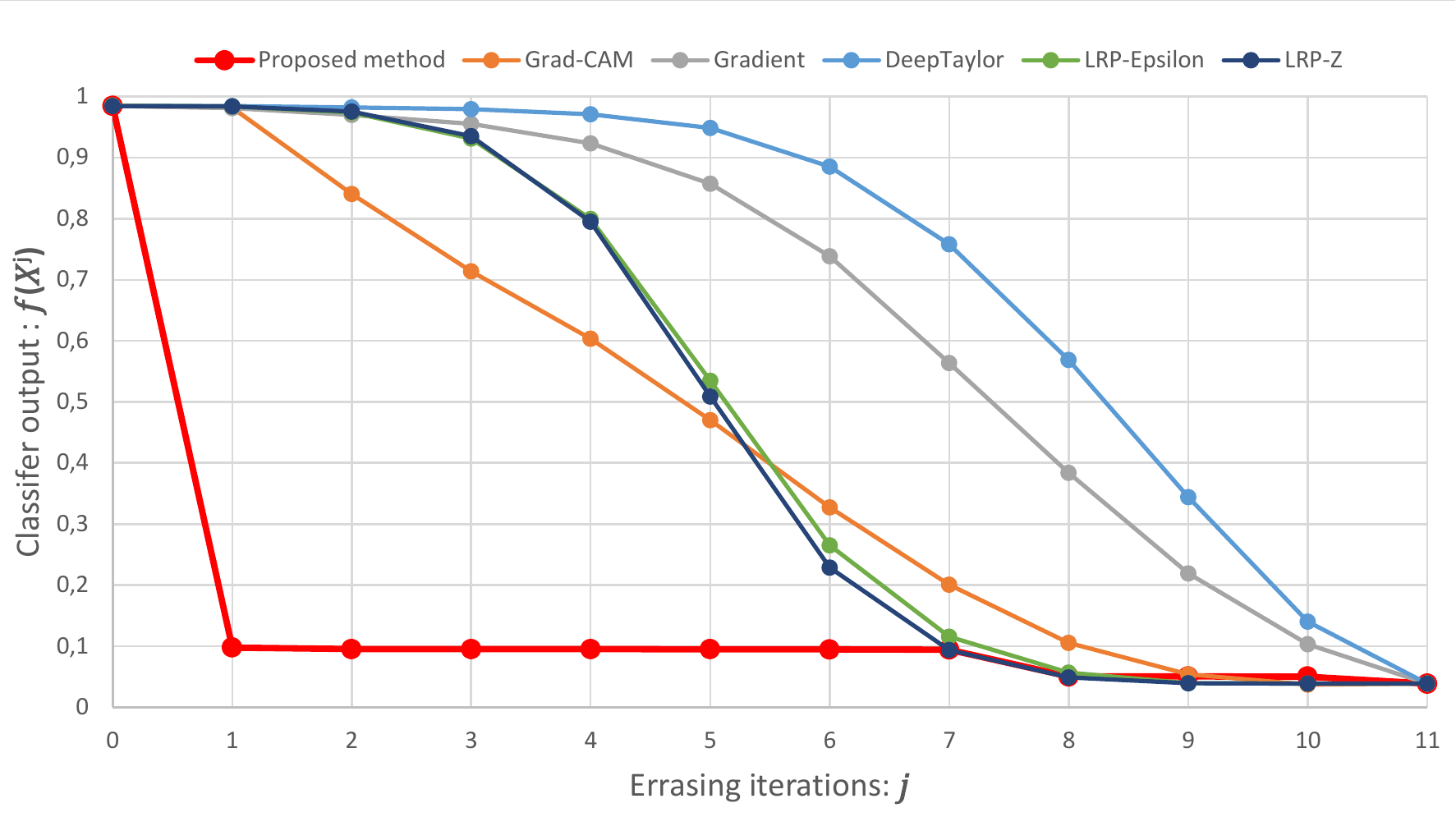}}
\caption{Perturbation curves of the visualization algorithms.}
\label{fig:comparison}
\end{figure}

\begin{equation}\label{eq:AOPC}
    AOPC = \frac{1}{11}\sum_{i=1}^{N}\sum_{j=1}^{11} \left ( f(X_i)-f(X_i^{(j)}) \right )
\end{equation}
\begin{table}[htbp]
  \centering
  \caption{Over the Perturbation Curve ($ AOPC $).}
    \begin{tabular}{|c|c|}
    \hline
    \multicolumn{1}{|p{9.10em}|}{\textbf{Proposed method}} & \textbf{0,907} \\
    \hline
    \textbf{Grad-CAM} & 0,587 \\
    \hline
    \textbf{Gradient} & 0,372 \\
    \hline
    \textbf{Deep Taylor} & 0,294 \\
    \hline
    \textbf{LRP-Z} & 0,558 \\
    \hline
    \textbf{LRP-Epsilon} & 0,550 \\
    \hline
    \end{tabular}%
  \label{tab:AOPC}%
\end{table}%
Fig.~\ref{fig:comparison} shows the perturbation curves of tested methods. The proposed method's curve decreases rapidly after erasing the pixels of the cluster $ C_3 $. Afterwards, before erasing pixels of the cluster $ C_2 $ the curve is stationary. Erasing pixels of the cluster $ C_2 $ decreases slightly $f$. Cluster $ C_1 $ does not contribute to the decreasing of the classifier output because this cluster contains background and non-discriminant pixels.  

Fig.~\ref{fig:comparison}  shows that the perturbation curves of the other methods decrease gradually because of the distribution of values in their heatmaps. On the other hand, the perturbation curve of the proposed method decreases steeply because the pixels of $ C_3 $ contains the most important pixels in respect to the classifier. 

Tab.~\ref{tab:AOPC} shows Area Over the Perturbation Curve ($ AOPC $) for each method. This quantity measures the decreases of the curve compared to its first value formulated in (\ref{eq:AOPC}) and Algorithm.1.%~\ref{Algo:1}. 

The proposed method has better $ AOPC = 0.907 $ than other methods  $ AOPC<0.6 $ because of the concentration of important pixels in $ C_3 $. In the proposed method, erasing only $ 13 \% $ of image  can decrease of $f$ to $0.1$. 

\section{Conclusion and further research}
In this work, we have proposed an interpretable Student/Teacher architecture for plant diseases classification. This architecture leverages the  multitask to produce a trainable visualization method. Our experiments demonstrate the benefit of adding the Student classifier to guide the architecture in order to reconstruct a sharp visualization images. This reconstructed images contain the discriminant regions, which helps to explain the classifier's decision. In the future, our objective is to test the Student/Teacher architecture on other classification problems. Besides, we will work to optimize the computation cost of this architecture.

\bibliographystyle{unsrt}  
\bibliography{./bibliography/my_bibl.bib}

\begin{thebibliography}{10}

\bibitem{HANSSEN201231}
Inge~M. Hanssen and Moshe Lapidot.
\newblock Chapter 2 - major tomato viruses in the mediterranean basin.
\newblock In Gad Loebenstein and Hervé Lecoq, editors, {\em Viruses and Virus
  Diseases of Vegetables in the Mediterranean Basin}, volume~84 of {\em
  Advances in Virus Research}, pages 31 -- 66. Academic Press, 2012.

\bibitem{Brahimi:2017}
Mohammed Brahimi, Kamel Boukhalfa, and Abdelouahab Moussaoui.
\newblock Deep learning for tomato diseases: Classification and symptoms
  visualization.
\newblock {\em Appl. Artif. Intell.}, 31(4):299--315, April 2017.

\bibitem{Fujita:2016}
E.~{Fujita}, Y.~{Kawasaki}, H.~{Uga}, S.~{Kagiwada}, and H.~{Iyatomi}.
\newblock Basic investigation on a robust and practical plant diagnostic
  system.
\newblock In {\em 2016 15th IEEE International Conference on Machine Learning
  and Applications (ICMLA)}, pages 989--992, Dec 2016.

\bibitem{Kawasaki:2015}
Yusuke Kawasaki, Hiroyuki Uga, Satoshi Kagiwada, and Hitoshi Iyatomi.
\newblock Basic study of automated diagnosis of viral plant diseases using
  convolutional neural networks.
\newblock In George Bebis, Richard Boyle, Bahram Parvin, Darko Koracin, Ioannis
  Pavlidis, Rogerio Feris, Tim McGraw, Mark Elendt, Regis Kopper, Eric Ragan,
  Zhao Ye, and Gunther Weber, editors, {\em Advances in Visual Computing},
  pages 638--645, Cham, 2015. Springer International Publishing.

\bibitem{Nachtigall:2016}
L.~G. {Nachtigall}, R.~M. {Araujo}, and G.~R. {Nachtigall}.
\newblock Classification of apple tree disorders using convolutional neural
  networks.
\newblock In {\em 2016 IEEE 28th International Conference on Tools with
  Artificial Intelligence (ICTAI)}, pages 472--476, Nov 2016.

\bibitem{Zeiler:13}
Matthew~D. Zeiler and Rob Fergus.
\newblock Visualizing and understanding convolutional networks.
\newblock {\em CoRR}, abs/1311.2901, 2013.

\bibitem{Springenberg:2015}
Jost~Tobias Springenberg, Alexey Dosovitskiy, Thomas Brox, and Martin~A.
  Riedmiller.
\newblock Striving for simplicity: The all convolutional net.
\newblock In {\em {ICLR} (Workshop)}, 2015.

\bibitem{Bach:2015}
Sebastian Bach, Alexander Binder, Grégoire Montavon, Frederick Klauschen,
  Klaus-Robert Müller, and Wojciech Samek.
\newblock On pixel-wise explanations for non-linear classifier decisions by
  layer-wise relevance propagation.
\newblock {\em PLOS ONE}, 10(7):1--46, 07 2015.

\bibitem{MONTAVON:2017}
Grégoire Montavon, Sebastian Lapuschkin, Alexander Binder, Wojciech Samek, and
  Klaus-Robert Müller.
\newblock Explaining nonlinear classification decisions with deep taylor
  decomposition.
\newblock {\em Pattern Recognition}, 65:211 -- 222, 2017.

\bibitem{Selvaraju:2017}
R.~R. {Selvaraju}, M.~{Cogswell}, A.~{Das}, R.~{Vedantam}, D.~{Parikh}, and
  D.~{Batra}.
\newblock Grad-cam: Visual explanations from deep networks via gradient-based
  localization.
\newblock In {\em 2017 IEEE International Conference on Computer Vision
  (ICCV)}, pages 618--626, Oct 2017.

\bibitem{Ronneberger:2015}
Olaf Ronneberger, Philipp Fischer, and Thomas Brox.
\newblock U-net: Convolutional networks for biomedical image segmentation.
\newblock In Nassir Navab, Joachim Hornegger, William~M. Wells, and
  Alejandro~F. Frangi, editors, {\em Medical Image Computing and
  Computer-Assisted Intervention -- MICCAI 2015}, pages 234--241, Cham, 2015.
  Springer International Publishing.

\bibitem{Simonyan:2015}
K.~Simonyan and A.~Zisserman.
\newblock Very deep convolutional networks for large-scale image recognition.
\newblock In {\em International Conference on Learning Representations}, 2015.

\bibitem{David:2015}
David~P. Hughes and Marcel Salath{\'{e}}.
\newblock An open access repository of images on plant health to enable the
  development of mobile disease diagnostics through machine learning and
  crowdsourcing.
\newblock {\em CoRR}, abs/1511.08060, 2015.

\bibitem{Simonyan:2014}
Karen Simonyan:2014, Andrea Vedaldi, and Andrew Zisserman.
\newblock Deep inside convolutional networks: Visualising image classification
  models and saliency maps.
\newblock In {\em {ICLR} (Workshop)}, 2014.

\end{thebibliography}
\end{document}